\documentclass[conference]{IEEEtran}
\IEEEoverridecommandlockouts
\usepackage{cite}
\usepackage{amsmath,amssymb,amsfonts}
\usepackage{algorithmic}
\usepackage{graphicx}
\usepackage{textcomp}
\usepackage{soul}
\usepackage{color,xcolor}
\usepackage{algorithm}
\usepackage{algorithmic}
\usepackage{subcaption}
\usepackage{booktabs}
\usepackage{multirow}
\def\BibTeX{{\rm B\kern-.05em{\sc i\kern-.025em b}\kern-.08em
    T\kern-.1667em\lower.7ex\hbox{E}\kern-.125emX}}
\begin{document}

\title{DELAYED RANDOM PARTIAL GRADIENT AVERAGING FOR FEDERATED LEARNING}

\author{\IEEEauthorblockN{Xinyi Hu}
\IEEEauthorblockA{\textit{ College of Information and Electronic Engineering} \\
\textit{Zhejiang University}\\
Hangzhou, China \\
Xinyih@zju.edu.cn}
}

\maketitle

\begin{abstract}
Federated learning (FL) is a distributed machine learning paradigm that enables multiple clients to train a shared model collaboratively while preserving privacy. 
However, the scaling of real-world FL systems is often limited by two communication bottlenecks: 
(a) while the increasing computing power of edge devices enables the deployment of large-scale Deep Neural Networks (DNNs), the limited bandwidth constraints frequent transmissions over large DNNs; and (b) high latency cost greatly degrades the performance of FL.  
In light of these bottlenecks, we propose a \textbf{D}elayed Random \textbf{P}artial \textbf{G}radient \textbf{A}veraging (DPGA) to enhance FL.
Under DPGA, clients only share partial local model gradients with the server. 
The size of the shared part in a local model is determined by the update rate, which is coarsely initialized and subsequently refined over the temporal dimension.
Moreover, DPGA largely reduces the system run time by enabling computation in parallel with communication.
We conduct experiments on non-IID CIFAR-10/100 to demonstrate the efficacy of our method.
\end{abstract}


\section{Introduction}
Federated learning (FL)~\cite{mcmahan2017communication,li2020federated, park2019wireless} is a privacy-preserving machine learning paradigm that enables multiple clients to jointly train a global model without sharing their local data.
In  contrast to conventional machine learning approaches that run on a central server, FL executes the training process on edge devices, where the intermediate parameters (e.g., weights or gradients) are  exchanged between the central server and the clients. 
Although the ever-increasing computational capacities of edge devices make it possible to deploy a large Deep Neural Network at the edge devices, the communication bottleneck, including the limited bandwidth and severe latency incurred by the spectral scarcity and long distances connection between clients and central server, constraints the scalability of FL systems.
In response, numerous methods~\cite{li2021fedmask,zhu2021delayed,chen2019communication, yang2022server} have been proposed to address this crucial issue in FL from two perspectives.

Approaches~\cite{chen2019communication,mohtashami2022masked} aim to improve communication efficiency via reducing the number of parameters uploaded per client to accommodate bandwidth limitations. 
For example, LG-Fed~\cite{liang2020think} only uploads the deep layer parameters, CD$^2$-pFed~\cite{shen2022cd2} decouples the channels and only shares part of channel parameters, and gradient compression~\cite{lin2017deep} upload partial gradient parameters.
However, once the local parameters are uploaded to the central server, clients need to wait for the results before they can proceed to the next round of local computations, as Fig.~\ref{fig:overview}.(a) shows, in these approaches with high network latency still. 
To reduce the system run time, a few approaches are proposed to take into account of the latency, such as DGA~\cite{zhu2021delayed} delays the gradient averaging step and allows local computation in parallel to communication, zero-wait SFWFL~\cite{yang2022server} allows clients to continuously perform local computing without being interrupted by the global parameter uploading on a server free wireless FL scenario, while these approaches all upload full model parameters as Fig.~\ref{fig:overview}.(b) shown.
It is worth noting that there is another asynchronous~\cite{xu2021asynchronous} FL approach that seems to solve a similar problem of latency, but what they actually solve is the heterogeneity between fast and slow clients, such as DAve-QN~\cite{soori2020dave} propose a distributed asynchronous averaging scheme of decision vectors and gradients in a way to effectively capture the local Hessian information of the objective function.
In summary, the above methods only address one aspect of the communication limitation in terms of bandwidth or latency, resulting in the communication of FL will still be constrained.

In this paper, we propose \textbf{D}elayed Random \textbf{P}artial \textbf{G}radient \textbf{A}veraging (DPGA) to simultaneously overcome the latency and bandwidth bottlenecks. 
On the one hand, to address the bandwidth limitation issue, we introduce an update rate to determine the number of uploaded parameters.
Recognizing that during different communication rounds, the global model has a different influence on the clients’ local computing performance, we employ the dynamic update rate based on the random walk process~\cite{lawler2010random} to accelerate the global training process while concurrently reducing the overall communication overhead.
It is worth noting that the architecture of the weights/gradient of client uploads varies in each communication round.
On top of the above, we proposed delayed partial gradient averaging so that clients continue their local computing during partial gradient communication and use these extra calculations to reduce the system run time as Fig.\ref{fig:overview}.(c) shown.
We evaluate our approach on CIFAR-10/100 datasets with non-IID settings. 
The results validate that our DPGA consistently outperforms the state-of-the-art methods in accuracy, communication cost, and run time.

\section{Proposed Method}
\label{sec:method}

\subsection{System model}
\label{sec:system model}
Consider an FL system consisting of a server and $N$ clients, in which client $i$ holds a local loss function $f_i(\cdot):\mathbb{R}^d\rightarrow\mathbb{R}$ constructed from its local dataset $\mathcal{D}_i$. 
The objective of all the entities in this system is to find a global model $w\in \mathbb{R}^d$ that solves the minimization problem 
\begin{equation}
    \min_{w} f(w)=\sum_i \frac{n_i}{n} f_i(w)
\end{equation}
where $n_i=|\mathcal{D}_i|$ denotes number of local data samples of client $i$ and $n=\sum_{j=1}^N|\mathcal{D}_j|$ is the total number of training samples across the system. 
In a typical communication round $t$, a subset of clients $\mathcal{S}_t$ are selected to conduct local training based on the latest global model weights $w_g^{t}$. 
Let $w_i^t$ denote the weights of client $i$'s model after local training. 
At the end of communication round $t$, the server would collect local models from the selected clients to update the global model via Federated Averaging (FedAvg~\cite{mcmahan2017communication}), i.e., $w_g^{t+1}= \sum_{i \in \mathcal{S}_t} \beta_i^t w_i^t$, in which $\beta_i^t= |\mathcal{D}_i|/\sum_{j \in \mathcal{S}_t}|\mathcal{D}_j|$ represents the ratio of the local data samples in client $i$ over the total number of data samples in the selected subset in communication round $t$.


In the partial gradient averaging paradigm of personalized federated learning, 
each client's gradient contains the personal and shared parts, although only the shared gradient will be exchanged between the clients and the global server. 
The ratio of the shared gradient to the total model gradient is defined as the \textit{update rate}:
\begin{equation}
    p^t = \frac{|\mathcal{G}^t_s|}{|\mathcal{G}^t_s| + |\mathcal{G}^t_p|},
\end{equation}
where $|\mathcal{G}^t_s|$ and $|\mathcal{G}^t_p|$ denote the amount of shared and personal gradient in the communication round $t$, respectively.

\subsection{Partial gradient averaging with dynamic update rates}
\begin{figure}[t]
\centering\includegraphics[width=0.5\textwidth]{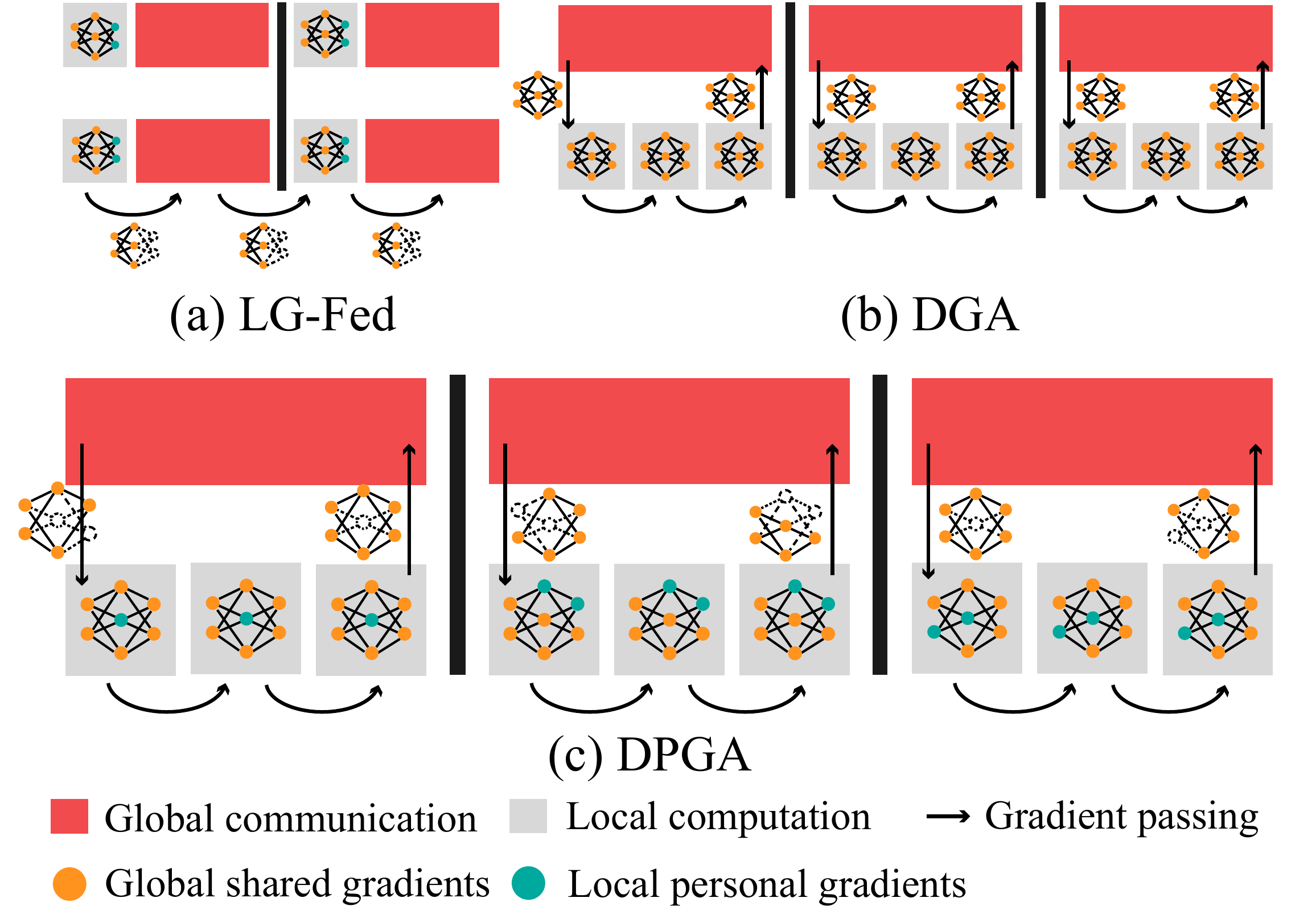} 
\caption{(a) Local computing and global updating are conducted sequentially with partial gradient passing. (b) Local computing is executed in parallel to global communication with full gradient exchange between the server and the clients. (c) Local computation is performed in parallel with global communication, and only partial gradient is exchanged.}

\label{fig:overview}
\end{figure}
We assign dynamic update rates $p \in (0,1]$ for local models based on the random walk process to renew the portion of the personal and shared gradient periodically. 

\subsubsection{Sample update rates based on the random walks}
A one-dimension random walk in discrete time is comprised of a walker that flips a fair coin and moves one step to the right or the left, depending on the result of the coin toss \cite{lawler2010random}. 
In consequence, the direction taken by the walker at each step is independent of the direction of the previous ones \cite{gonzalez2022random}. The walker's position after $m$ steps can be expressed as 
\begin{equation}
    x' = x + X_1 + \dots + X_m,
\end{equation}
where $x$ is the initial position and $X_i \in \{ -1, 1\}$ is a discrete random variable. 

Apart from the probabilistic perspective, a one-dimension random walk can also be modeled by a Markov chain~\cite{geyer1992practical} to represent the stochastic process of sampling update rates.
To be more concrete, in communication round $t$, we have a state $Q^t(p^t)$, denoting that the update rate equals $p^t$. 
Then, the update rate $p^{t+1}$ in the $t+1$-th communication round is obtained by an $m$-step random walk with starting point $p^t$ (the step size is set to 0.1). The transition probability from state $Q^t(p^t)$ to state $Q^{t+1} (p^{t+1})$ is given by
\begin{equation}
    \phi_t(p^{t+1}) = \frac{C_m^{A\vert p^{t+1} - p^t \vert}}{2^m},
    \label{transition}
\end{equation}
where $C_x^y = \frac{y!}{x!(y-x)!}$, $A = 10$. 
Equation~(\ref{transition}) indicates that the larger the gap between $p^t$ and $p^{t+1}$, the smaller the transition probability. Moreover, the transition probabilities are normalized as
\begin{equation}
    \sum_{p^{t+1}} \phi_t(p^{t+1}) = 1.
\end{equation}

\begin{algorithm}[tb]
\caption{Delayed Partial Gradient Averaging (DPGA).}
\label{alg:DPGA}
\textbf{Input}: Rounds $T$, local epoch $K$, communication latency represented in the number of computing rounds $D$.\\
\textbf{Output}: local model of each client 
\begin{algorithmic}[1] 
\FOR{each round $t$ from 1 to $T$}
\STATE Sample update rate $p^t$ based on the random walk.
\FOR{each client in parallel}
\FOR{each local epoch $k$ from 1 to $K$}
\STATE When the delayed global gradient is received, replace the shared part of the corresponding local gradient with it.
\STATE update the full local model $w_{n}^{(t+1)}$ by forward and backward propagation. \\ 
\begin{small}
    \centering
    \begin{equation}
    \nonumber
        w_i^{t+D,k+1}= 
      \begin{cases}
      w_i^{t+D,k} - \eta g_i^{t+D,k} &, \mbox{if $t$ mod $D \neq 1$} \\
      & \mbox{or $k \neq 1$,}\\
      w^{t+D,k}_i - \eta z_i^{t+D} - &\\
      \eta(-z_i^{t}(p^{t}) + \overline{g^{t}}(p^{t}_i)) &, \mbox{otherwise.}
      \end{cases}
      \end{equation}
\end{small}

\ENDFOR

\STATE Send the $t$-th communication round global shared gradient $\overline{g^t}(p^t)=\frac{1}{N}\sum_{i=1}^{N}{g^t_i(p^t)}$ to all clients.
\ENDFOR
\ENDFOR
\end{algorithmic}
\end{algorithm}

\subsubsection{Partial gradient averaging}
Recall that the server and clients only exchange global shared gradient during the partial gradient averaging process. 

Due to the diverse architectures of the received partial gradient, the server would update the global model via component-wise aggregation. 
For each component (i.e., layer, channel, weight, etc.), the global gradient $\overline{g^t}$ are updated over the subset of clients whose architecture contains the corresponding component.
For communication round $t$, client $i$ downloads the partial global gradient $\overline{g^t}(p_i^t)$ and merges it with the partial local gradient $g^t_i(1-p_i^t)$. 
The merged result is given by
\begin{equation}
    g^t_i = \overline{g^t}(p_i^t) \cup g^t_i(1-p_i^t) \\
          = \overline{g^t} + g^t_i(1-p_i^t),
\end{equation}
The expression of a typical partial gradient is given as
\begin{equation}
    g^t(p^t) = g^t \odot m(p^t),
\end{equation}
in which $g^t_i$ is the gradient of the full model, $m(p_i^t)$ denotes the local personal mask of the client $i$. 
The local personal and global shared masks are complementary. We determined the location of the upload gradient parameters, whose local personal mask is set to 0, by Top-K sparsification~\cite{shi2019understanding}.
And the dynamic ‘K’ determined by the update rate $p^t$. 
Then the rest of the global shared mask is set to 1.

\subsection{Delayed random partial gradient averaging}

Consider of the inevitable high communication latency of the gradient averaging, we propose the DPGA, which allows local updates during the download/upload partial gradient, to reduce the system run time.
Specifically, the global gradient $\overline{g^t}$ is delayed to a later iteration so that clients can immediately start the next round and a partial gradient correction term is designed to compensate the staleness.

To simplify the notation representation, we assume that all clients spend the same amount of time uploading and downloading partial gradients in each round of global communication.
The communication time spreads over $D$ local computing rounds, which is in total $tD$ iterations.
Then, as shown in Fig.~\ref{fig:overview}.(c), the DPGA no longer freezes the local computation power during the communication.
To make the discussion more explicit, we denote the model weights on the $i$-th client at the $k$-th iteration within the $t$-th round by $w^{t,k}_i$, and the corresponding stochastic gradient as $g^{t,k}_i$.
In the first round of local computing, each client $i$ executes $K$ stochastic gradient descent (SGD) iterations and arrives at the following:
\begin{equation}
\begin{aligned}
    w^{1,K+1}_i &= w^{1,K}_i - \eta \nabla f_i(w^{1,K}_i)\\
                &= \cdots = w_0 - \eta \sum_{k=1}^K \nabla f_i(w^{1,k}_i) \\
                &= w_0 - \eta \underbrace{\sum_{k=1}^K g_i^{1,k}}_{:=z^1_i},
\end{aligned}
\end{equation}
where, $w_0$ is the initial weights of the local models.
With the end of the first round's computation, the accumulated partial gradient $z^1_i:=\sum_{k=1}^K g_i^{1,k}(p^1)$ is sent to all the clients, i.e. partial gradient averaging.
Then, we immediately execute the second round's local updates, leaving the first round averages in transmission.
When the global partial gradient is received, client $i$ has already performed D extra rounds of local updates and its model weights can be written as:
\begin{equation}
    \begin{aligned}
        w^{1+D,1}_i &= w^{D,K}_i - \eta \nabla f_i(w^{D,K}_i)\\
        &= w_i^{D,1} - \eta z_i^D \\
        &=w_0 - \eta(z_i^D + z_i^{D-1}+ \cdots +z_i^1).
    \end{aligned}
\end{equation}
As the global gradient $\overline{g^1}$, of the first round is now available, client $i$ can substitute the shared part of the first round local gradients by the global one
\begin{equation}
\begin{aligned}
    w^{1+D,1}_i &= (w_0 - \eta z_i^1) - \eta (z_i^2 +  \cdots + z_i^D) \\
    &= w^{D,1}_i -\eta(z_i^D - z_i^1(p_i^1)+ \overline{g^1}(p_i^1)),
\end{aligned}
\end{equation}
where, $z_i^t(p_i^t) = \sum_{k=1}^Kg_i^{t,k}(p_i^t)$ represents the global shared part gradient of the client $i$ in the $t$-th communication round.

The details of DPGA is elucidated in Algorithm~\ref{alg:DPGA}.
This algorithm contains two major component: 
(a) The partial gradient communication with dynamic update rates, which breakthrough the bandwidth bottleneck, and (b) Delayed gradient aggregation, which allows local computation in parallel to communication.
A comparison among the sequential training procedure where computations are suspended during global communications, parallel training procedure and our DPGA is provided in Fig.~\ref{fig:overview}. 

\section{Numerical Results}
\label{sec:exp}
In this section, we conduct simulations to evaluate the performance of DPGA on a LeNet-5~\cite{lecun1998gradient} over the CIFAR-10~\cite{krizhevsky2009learning} and on a ResNet-34 ~\cite{he2016deep} over the CIFAR-100~\cite{krizhevsky2009learning} with IID and non-IID data heterogeneity settings.
We adopt Dirichlet distribution-based non-IID data partition method~\cite{xu2022fedcorr} for CIFAR-10/100, where the identicalness of local data distribution and the class imbalance could be controlled by the parameter pair $(\alpha, \rho)$.
We compare our DPGA with conventional FL method (FedAvg), partial gradient averaging (LG-Fed) methods, and delayed gradient averaging (DGA) methods from three dimensions, i.e., test accuracy, communication time, and communication parameters.
The total number of bytes transmitted through uplink (clients-to-server) and downlink (server-to-clients) connections represents the communication cost throughout the training. 
Also, we express the communication time as a ratio of the communication to the local computation time. 
Since the FedAvg, LG-Fed, DGA, and DPGA are trained locally with the full model, it can be assumed that the local computation time per round is the same for these four methods.
All simulations are conducted on NVIDIA TITAN Xp GPU.

\begin{table}[]
\renewcommand\arraystretch{2}
\setlength{\tabcolsep}{4.5mm}
\centering
\caption{A comparison of test accuracy (\%) on CIFAR-10.
}

\begin{tabular}{ccccc}

\toprule
& \multirow{2}{*}{IID}&  \multicolumn{3}{c}{non-IID (Dirichlet distribution)} \\ \cline{3-5}     
&  &  $(10, 0.7)$ & $(1, 1)$ &  $ (0.1, 1)$
\\ \hline
FedAvg & 60.08 & 63.62 & 44.76 & 56.32 \\ 
\hline
DGA & 61.02 & 62.05 & 46.01 & 55.32 \\ 
 \hline
LG-Fed & 62.14 & 69.01 & 46.21 & 56.97 \\ 
 \hline
DPGA & {\textbf{93.08}} & {\textbf{81.1}} & {\textbf{69.96}} & {\textbf{86.01}}  
\\
\bottomrule
\end{tabular}
\label{tab:cifar10}
\end{table}

First, we compare the performance of four different federated learning methods on CIFAR-10.
Table \ref{tab:cifar10} shows the results including four different settings with varying degrees of data heterogeneity. 
The first setting is IID, followed by the three non-IID settings based on the Dirichlet distribution partitioning $(\alpha, \rho) = (10,0.7), (1,1), (0.1,1)$, whose degree of data heterogeneity increases sequentially.
In the above setting, the number of clients $N$ is 1000, 1000, 500, and 100, respectively.
The results showed that DPGA outperformed the other methods in all settings, achieving the highest accuracy. 
For example, in the IID setting, DPGA achieved a test accuracy of 93.08\%, while the other methods FedAvg, DGA, and LG-Fed achieved 60.08\%, 61.02\%, and 62.14\%, respectively. 
As the degree of data heterogeneity increases, such as at $(\alpha, \rho) = (0.1,1)$, DPGA still maintains a high test accuracy (86.01\%) while the test accuracy of other methods does not exceed 60\%.

\begin{table}[]
\renewcommand\arraystretch{2}
\setlength{\tabcolsep}{4.5mm}
\centering
\caption{A comparison of test accuracy (\%) on CIFAR-100.
}

\begin{tabular}{ccccc}

\toprule
& \multirow{2}{*}{IID}&  \multicolumn{3}{c}{non-IID (Dirichlet distribution)} \\ \cline{3-5}     
&  &  $(10, 0.7)$ & $(1, 1)$ &  $ (0.1, 1)$
\\ \hline
FedAvg & 51.03 & 40.27 & 43.08 & 43.95 \\ 
\hline
DGA & 50.59 & 40.03 & {43.10} & {43.81} \\ 
 \hline
LG-Fed & {53.21} & {50.06} & {54.29} & {50.39} \\ 
 \hline
DPGA & {\textbf{71.22}} & {\textbf{69.07}} & {\textbf{68.33}} & {\textbf{80.06}}  
\\
\bottomrule
\end{tabular}
\label{tab:cifar100}
\end{table}

To provide a comprehensive evaluation, we also compared our method with the other methods on CIFAR-100. 
The same as CIFAR-10, we conducted simulations on four different data heterogeneity settings, including IID, three settings based on Dirichlet distribution partitioning $(\alpha, \rho) = (10,0.7), (1,1), (0.1,1)$.
The whole CIFAR-100 dataset is participated in 100 non-overlapped portions and assigned to the $N=100$ clients in all settings.
Table ~\ref{tab:cifar100} shows that our DPGA still outperforms other methods in all data heterogeneity settings, achieving the highest accuracy.
Specifically, in the highly non-IID setting $(\alpha,\rho)=(0.1,1)$, the DPGA achieves an excellent accuracy of 80.06\%, which is significantly higher than the second-best performing algorithm, LG-Fed, with an accuracy of 50.39\%.

\begin{figure}[htb]

\begin{minipage}[b]{.48\linewidth}
  \centering
  \centerline{\includegraphics[width=4.8cm]{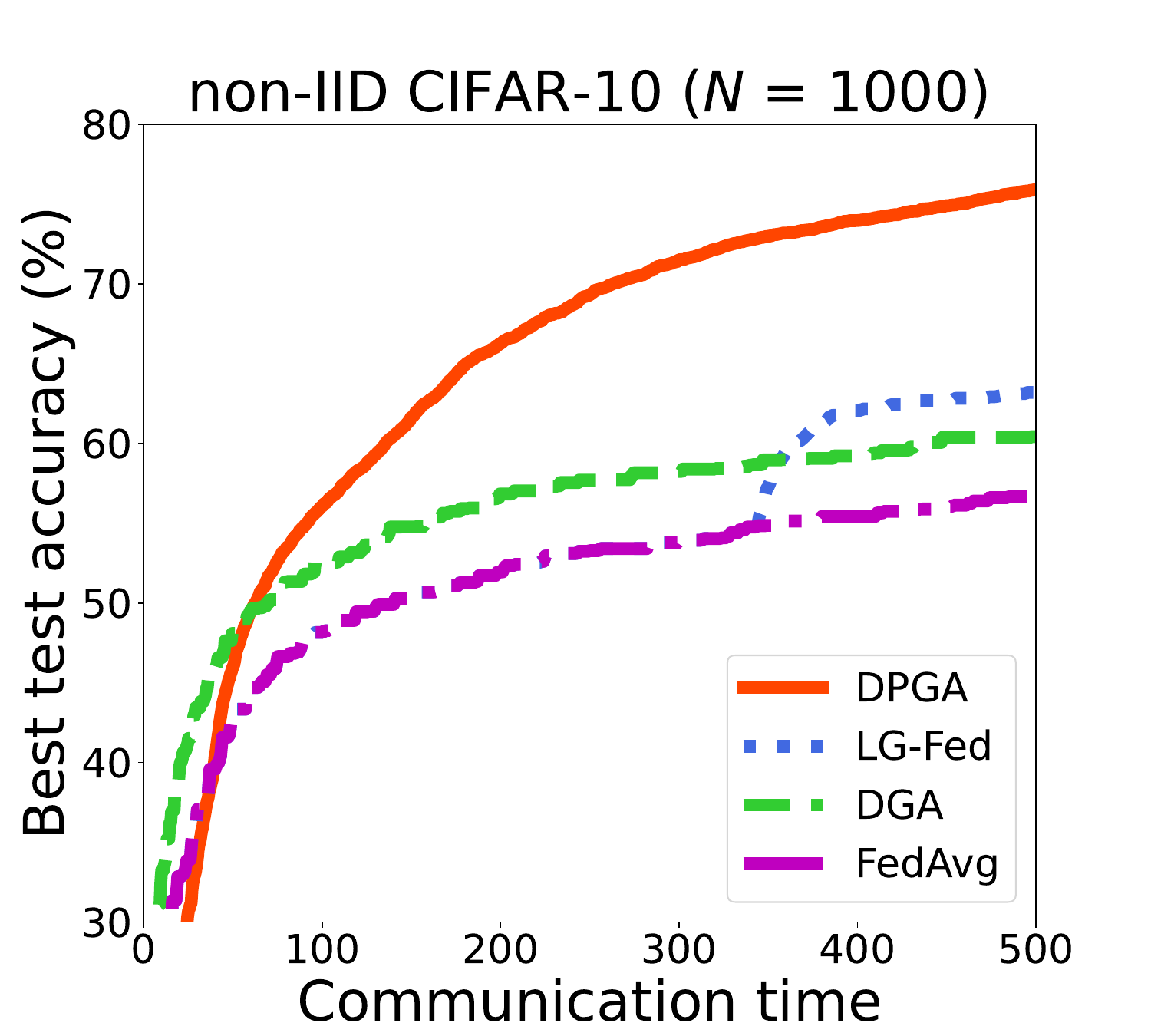}}
  \centerline{(a)}\medskip
\end{minipage}
\hfill
\begin{minipage}[b]{0.48\linewidth}
  \centering
  \centerline{\includegraphics[width=4.8cm]{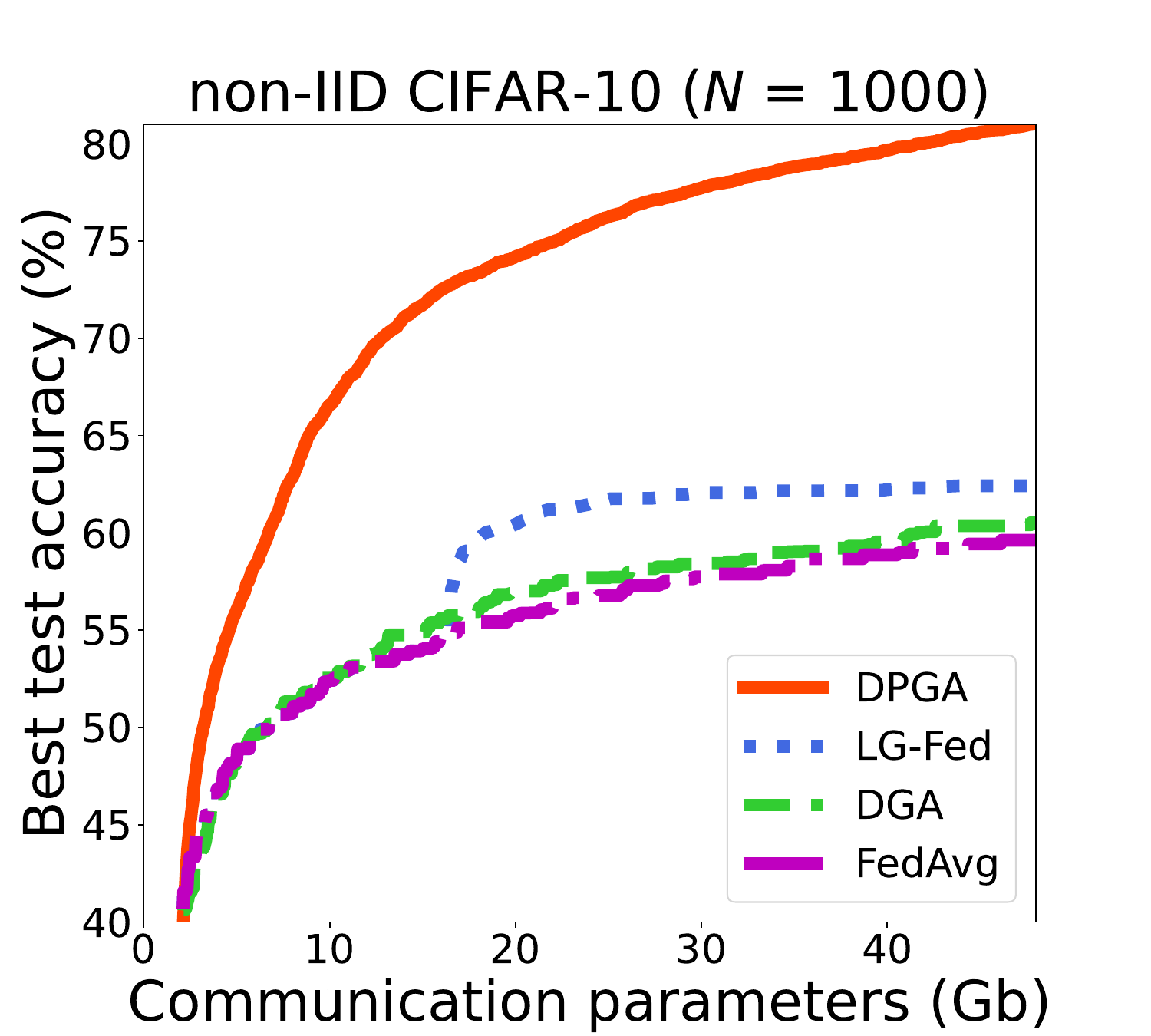}}
  \centerline{(b)}\medskip
\end{minipage}
\begin{minipage}[b]{.48\linewidth}
  \centering
  \centerline{\includegraphics[width=4.8cm]{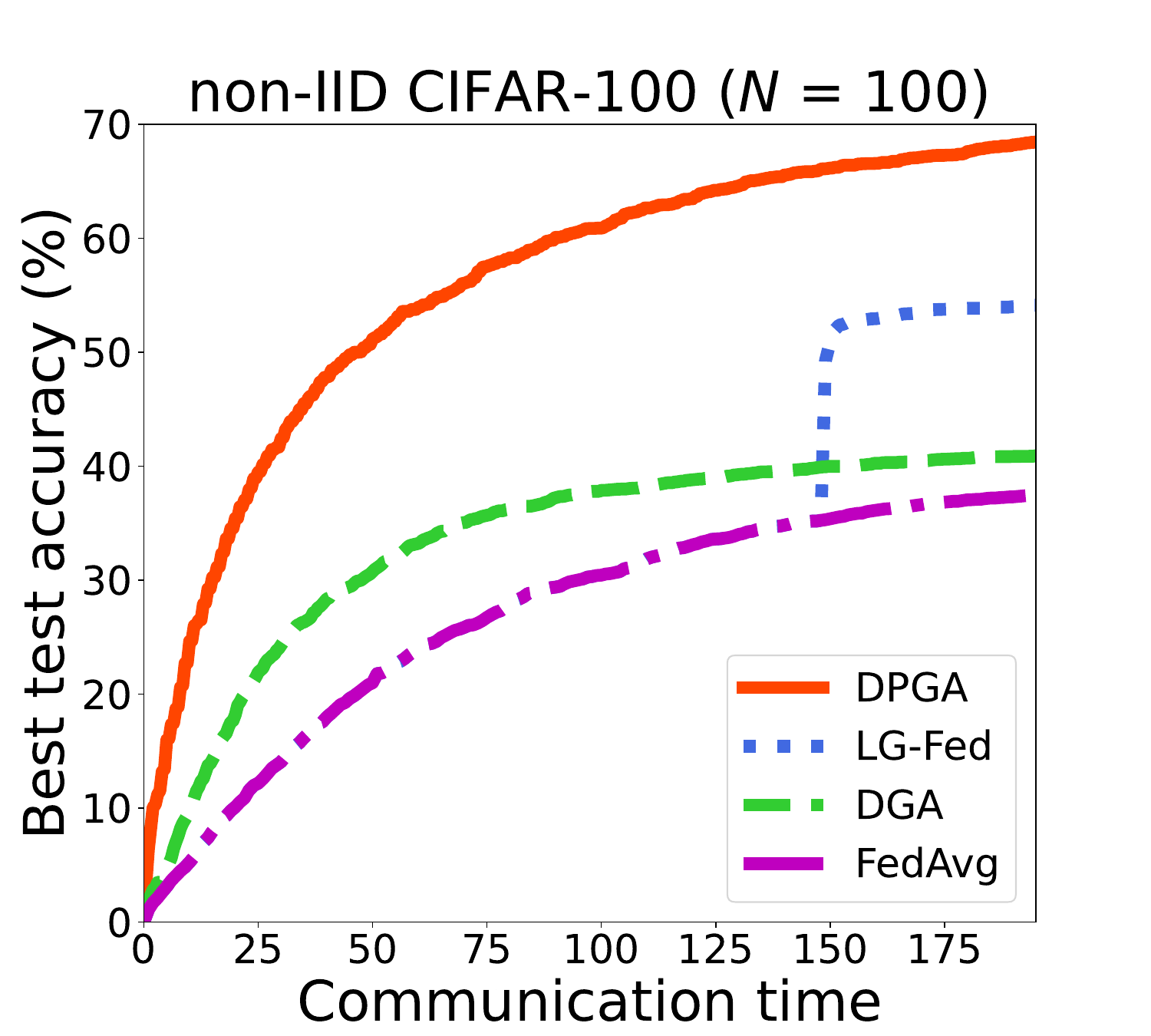}}
  \centerline{(c)}\medskip
\end{minipage}
\hfill
\begin{minipage}[b]{0.48\linewidth}
  \centering
  \centerline{\includegraphics[width=4.8cm]{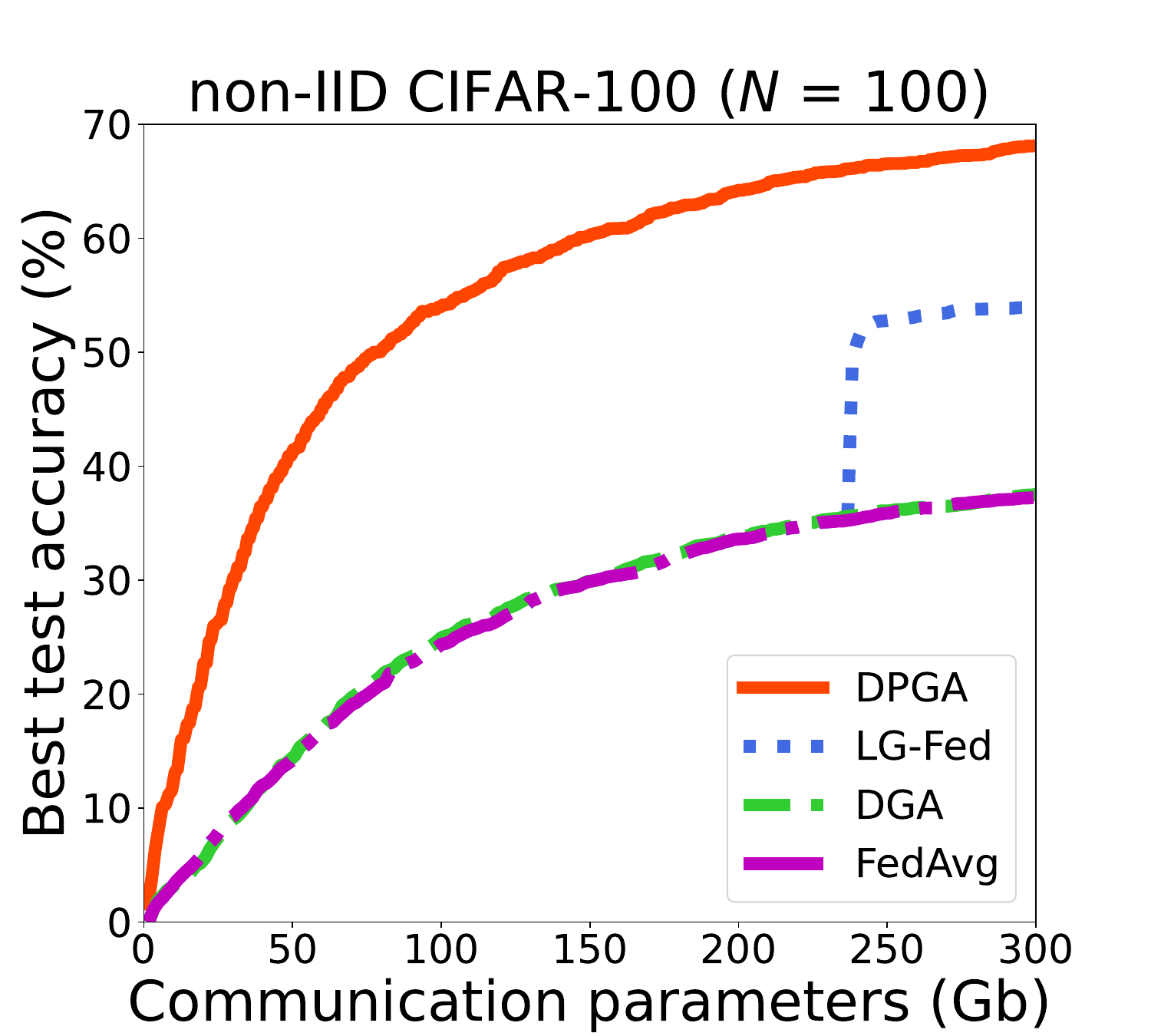}}
  \centerline{(d)}\medskip
\end{minipage}
\caption{A comparison of different methods with non-IID CIFAR-10/100 settings. LG-Fed is fine-tuned based on the FedAvg pre-trained model.}
\label{fig:visual}
\end{figure}

Moreover, our DPGA significantly improves communication efficiency.
Figure~\ref{fig:visual} visualize the training process of the four methods using the non-IID CIFAR-10/100 setting ($(\alpha,\rho)=(10,0.7)$) as an example.
Figure~\ref{fig:visual}~(a)\&(b) show the variation of test accuracy with communication time and parameters under the CIFAR-10 dataset. 
To achieve 60\% test accuracy, the required communication time of DPGA is only 150 communication time, while LG-Fed and DGA require about 500, and FedAvg requires more. 
In addition, DPGA requires only 6.7 Gb of communication parameters, which is 4\% of FedAvg and DGA. Even the second-best performing LG-Fed requires three times the parameters of DPGA to achieve 60\% accuracy.
Figure~\ref{fig:visual}~(c)\&(d) visualize the training process under the CIFAR-100 dataset.
When achieving 40\% test accuracy, the communication time of DPGA is only 36, while FedAvg, DGA, and LG-Fed are 349, 234, and 152, respectively. 
From the perspective of communication parameters, our DPGA also has a much smaller overhead than the other three methods. 
Specifically, DPGA requires only 67.3 Gb of parameters to achieve the target accuracy, while the other three methods require communication parameters ranging from 247 Gb to 357 Gb.\looseness=-1

Compare to the conventional FL method (FedAvg), the method for solving latency problems (DGA) achieves the target accuracy in a shorter time. Since it exchanges the full model between clients and server, the communication parameters are not reduced.
And the method for solving bandwidth limitations (LG-Fed) achieves the target accuracy with fewer communication parameters but requires a longer communication time than DGA.
Our DPGA solves both latency and bandwidth bottlenecks and performs much better than DGA and LG-Fed in terms of communication time and communication parameters when achieving the target accuracy.
In addition, DPGA achieves much higher test accuracy in a variety of heterogeneous settings than the other methods.

Comprehensive simulations on different datasets demonstrate that our DPGA consistently outperforms those conventional FL methods with the extremely much less communication time and parameters.

\section{Conclusion}
In this paper, we proposed DPGA for FL. By adopting partial gradient exchange and the parallel strategy for communication and local computation, our scheme is able to break through both bandwidth and latency bottlenecks.
We evaluate our approach on CIFAR-10/100 datasets with non-IID settings. The results validate that our DPGA consistently outperforms the state-of-the-art methods in accuracy, communication cost, and run time.

\bibliographystyle{IEEEtran}
\bibliography{refs}

\begin{thebibliography}{10}
\providecommand{\url}[1]{#1}
\csname url@samestyle\endcsname
\providecommand{\newblock}{\relax}
\providecommand{\bibinfo}[2]{#2}
\providecommand{\BIBentrySTDinterwordspacing}{\spaceskip=0pt\relax}
\providecommand{\BIBentryALTinterwordstretchfactor}{4}
\providecommand{\BIBentryALTinterwordspacing}{\spaceskip=\fontdimen2\font plus
\BIBentryALTinterwordstretchfactor\fontdimen3\font minus
  \fontdimen4\font\relax}
\providecommand{\BIBforeignlanguage}[2]{{%
\expandafter\ifx\csname l@#1\endcsname\relax
\typeout{** WARNING: IEEEtran.bst: No hyphenation pattern has been}%
\typeout{** loaded for the language `#1'. Using the pattern for}%
\typeout{** the default language instead.}%
\else
\language=\csname l@#1\endcsname
\fi
#2}}
\providecommand{\BIBdecl}{\relax}
\BIBdecl

\bibitem{mcmahan2017communication}
B.~McMahan, E.~Moore, D.~Ramage, S.~Hampson, and B.~A. y~Arcas,
  ``Communication-efficient learning of deep networks from decentralized
  data,'' in \emph{Artificial intelligence and statistics}.\hskip 1em plus
  0.5em minus 0.4em\relax PMLR, 2017, pp. 1273--1282.

\bibitem{li2020federated}
T.~Li, A.~K. Sahu, M.~Zaheer, M.~Sanjabi, A.~Talwalkar, and V.~Smith,
  ``Federated optimization in heterogeneous networks,'' \emph{Proceedings of
  Machine Learning and Systems}, vol.~2, pp. 429--450, 2020.

\bibitem{park2019wireless}
J.~Park, S.~Samarakoon, M.~Bennis, and M.~Debbah, ``Wireless network
  intelligence at the edge,'' \emph{Proceedings of the IEEE}, vol. 107, no.~11,
  pp. 2204--2239, 2019.

\bibitem{li2021fedmask}
A.~Li, J.~Sun, X.~Zeng, M.~Zhang, H.~Li, and Y.~Chen, ``Fedmask: Joint
  computation and communication-efficient personalized federated learning via
  heterogeneous masking,'' in \emph{Proceedings of the 19th ACM Conference on
  Embedded Networked Sensor Systems}, 2021, pp. 42--55.

\bibitem{zhu2021delayed}
L.~Zhu, H.~Lin, Y.~Lu, Y.~Lin, and S.~Han, ``Delayed gradient averaging:
  Tolerate the communication latency for federated learning,'' \emph{Advances
  in Neural Information Processing Systems}, vol.~34, pp. 29\,995--30\,007,
  2021.

\bibitem{chen2019communication}
Y.~Chen, X.~Sun, and Y.~Jin, ``Communication-efficient federated deep learning
  with layerwise asynchronous model update and temporally weighted
  aggregation,'' \emph{IEEE transactions on neural networks and learning
  systems}, vol.~31, no.~10, pp. 4229--4238, 2019.

\bibitem{yang2022server}
H.~H. Yang, Z.~Chen, and T.~Q. Quek, ``Server free wireless federated learning:
  Architecture, algorithm, and analysis,'' \emph{arXiv preprint
  arXiv:2204.07609}, 2022.

\bibitem{mohtashami2022masked}
A.~Mohtashami, M.~Jaggi, and S.~Stich, ``Masked training of neural networks
  with partial gradients,'' in \emph{International Conference on Artificial
  Intelligence and Statistics}.\hskip 1em plus 0.5em minus 0.4em\relax PMLR,
  2022, pp. 5876--5890.

\bibitem{liang2020think}
P.~P. Liang, T.~Liu, L.~Ziyin, N.~B. Allen, R.~P. Auerbach, D.~Brent,
  R.~Salakhutdinov, and L.-P. Morency, ``Think locally, act globally: Federated
  learning with local and global representations,'' \emph{arXiv preprint
  arXiv:2001.01523}, 2020.

\bibitem{shen2022cd2}
Y.~Shen, Y.~Zhou, and L.~Yu, ``Cd2-pfed: Cyclic distillation-guided channel
  decoupling for model personalization in federated learning,'' in
  \emph{Proceedings of the IEEE/CVF Conference on Computer Vision and Pattern
  Recognition}, 2022, pp. 10\,041--10\,050.

\bibitem{lin2017deep}
Y.~Lin, S.~Han, H.~Mao, Y.~Wang, and W.~J. Dally, ``Deep gradient compression:
  Reducing the communication bandwidth for distributed training,'' in
  \emph{International Conference on Learning Representations}, 2017.

\bibitem{xu2021asynchronous}
C.~Xu, Y.~Qu, Y.~Xiang, and L.~Gao, ``Asynchronous federated learning on
  heterogeneous devices: A survey,'' \emph{arXiv preprint arXiv:2109.04269},
  2021.

\bibitem{soori2020dave}
S.~Soori, K.~Mishchenko, A.~Mokhtari, M.~M. Dehnavi, and M.~Gurbuzbalaban,
  ``Dave-qn: A distributed averaged quasi-newton method with local superlinear
  convergence rate,'' in \emph{International Conference on Artificial
  Intelligence and Statistics}.\hskip 1em plus 0.5em minus 0.4em\relax PMLR,
  2020, pp. 1965--1976.

\bibitem{lawler2010random}
G.~F. Lawler, \emph{Random walk and the heat equation}.\hskip 1em plus 0.5em
  minus 0.4em\relax American Mathematical Soc., 2010, vol.~55.

\bibitem{gonzalez2022random}
F.~H. Gonz{\'a}lez, ``Random walks on networks with stochastic reset to
  multiple nodes,'' \emph{arXiv preprint arXiv:2204.10885}, 2022.

\bibitem{geyer1992practical}
C.~J. Geyer, ``Practical markov chain monte carlo,'' \emph{Statistical
  science}, pp. 473--483, 1992.

\bibitem{shi2019understanding}
S.~Shi, X.~Chu, K.~C. Cheung, and S.~See, ``Understanding top-k sparsification
  in distributed deep learning,'' \emph{arXiv preprint arXiv:1911.08772}, 2019.

\bibitem{lecun1998gradient}
Y.~LeCun, L.~Bottou, Y.~Bengio, and P.~Haffner, ``Gradient-based learning
  applied to document recognition,'' \emph{Proceedings of the IEEE}, vol.~86,
  no.~11, pp. 2278--2324, 1998.

\bibitem{krizhevsky2009learning}
A.~Krizhevsky, G.~Hinton \emph{et~al.}, ``Learning multiple layers of features
  from tiny images,'' 2009.

\bibitem{he2016deep}
K.~He, X.~Zhang, S.~Ren, and J.~Sun, ``Deep residual learning for image
  recognition,'' in \emph{Proceedings of the IEEE conference on computer vision
  and pattern recognition}, 2016, pp. 770--778.

\bibitem{xu2022fedcorr}
J.~Xu, Z.~Chen, T.~Q. Quek, and K.~F.~E. Chong, ``Fedcorr: Multi-stage
  federated learning for label noise correction,'' in \emph{Proceedings of the
  IEEE/CVF Conference on Computer Vision and Pattern Recognition}, 2022, pp.
  10\,184--10\,193.

\end{thebibliography}
\end{document}